\documentclass{article}

\usepackage{graphicx}
\usepackage{hyperref}

\def\code#1{\texttt{#1}}
\usepackage{amsmath,amsfonts,stmaryrd,amssymb} 

\usepackage{enumerate} 

\usepackage[ruled]{algorithm2e} 

\usepackage[framemethod=tikz]{mdframed} 

\usepackage{listings} 
\usepackage{geometry} 

\usepackage[utf8]{inputenc} 
\usepackage[T1]{fontenc} 

\usepackage{XCharter} 

\mdfdefinestyle{commandline}{
	leftmargin=10pt,
	rightmargin=10pt,
	innerleftmargin=15pt,
	middlelinecolor=black!50!white,
	middlelinewidth=2pt,
	frametitlerule=false,
	backgroundcolor=black!5!white,
	frametitle={Command Line},
	frametitlefont={\normalfont\sffamily\color{white}\hspace{-1em}},
	frametitlebackgroundcolor=black!50!white,
	nobreak,
}

\mdfdefinestyle{file}{
	innertopmargin=1.6\baselineskip,
	innerbottommargin=0.8\baselineskip,
	topline=false, bottomline=false,
	leftline=false, rightline=false,
	leftmargin=2cm,
	rightmargin=2cm,
	singleextra={%
		\draw[fill=black!10!white](P)++(0,-1.2em)rectangle(P-|O);
		\node[anchor=north west]
		at(P-|O){\ttfamily\mdfilename};
		\def\l{3em}
		\draw(O-|P)++(-\l,0)--++(\l,\l)--(P)--(P-|O)--(O)--cycle;
		\draw(O-|P)++(-\l,0)--++(0,\l)--++(\l,0);
	},
	nobreak,
}

\mdfdefinestyle{question}{
	innertopmargin=1.2\baselineskip,
	innerbottommargin=0.8\baselineskip,
	roundcorner=5pt,
	nobreak,
	singleextra={%
		\draw(P-|O)node[xshift=1em,anchor=west,fill=white,draw,rounded corners=5pt]{%
		Question \theQuestion\questionTitle};
	},
}

\newcounter{Question} 

\mdfdefinestyle{warning}{
	topline=false, bottomline=false,
	leftline=false, rightline=false,
	nobreak,
	singleextra={%
		\draw(P-|O)++(-0.5em,0)node(tmp1){};
		\draw(P-|O)++(0.5em,0)node(tmp2){};
		\fill[black,rotate around={45:(P-|O)}](tmp1)rectangle(tmp2);
		\node at(P-|O){\color{white}\scriptsize\bf !};
		\draw[very thick](P-|O)++(0,-1em)--(O);
	}
}

\mdfdefinestyle{info}{%
	topline=false, bottomline=false,
	leftline=false, rightline=false,
	nobreak,
	singleextra={%
		\fill[black](P-|O)circle[radius=0.4em];
		\node at(P-|O){\color{white}\scriptsize\bf i};
		\draw[very thick](P-|O)++(0,-0.8em)--(O);
	}
}

\newenvironment{info}[1][Info:]{ 
	\medskip
	\begin{mdframed}[style=info]
		\noindent{\textbf{#1}}
}{
	\end{mdframed}
}

\title{ \textbf{CMPS 396X}\\
American University of Beirut \\
Data Science Term Project: \\
Mangrove Geographic Distribution Over Time} 

\author{Ezzat Chebaro, Jad Ismail, Amir Nasrelddine, Lynn Wahab, Ali El-Zein}

\date{\today}

\begin{document}

\maketitle 
\newpage

\tableofcontents
\newpage

\section{Pre-Stage: Project Idea}

Climate change is an impending disaster which is of pressing concern more and more every year. Countless efforts have been made to study the long-term effects of climate change on agriculture \cite{Sutariya}, land resources \cite{Mendelsohn}, and biodiversity \cite{Reed}. Studies involving marine life, however, are less prevalent in the literature. Inspired by the Netflix series ’Our Planet’, our research studies the available data on the population of mangroves (groups of shrubs or small trees living in saline coastal intertidal zones) and their correlations to climate change statistics, specifically, temperature, heat content, various sea levels, and sea salinity. Coastal seas, which make up only 7\% of the ocean, host approximately 95\% of the world’s marine life and all the world’s mangroves, thus making them essential to marine biodiversity \cite{FAO}. Mangroves are especially relevant to oceanic ecosystems because of their protective nature towards other marine life, as well as their high absorption rate of carbon dioxide, and their ability to withstand varying levels of salinity of our coasts. 

After a brief literature review, we identified factors that contribute to the decline of mangroves: temperature, salinity, and sea levels \cite{Carugati, Spalding}. The goal of the project is to use these factors to build a learning model that forecasts the geographic distribution of mangroves over the next year. If our models are accurate and powerful, in the next 5 years they could potentially influence crucial policy decisions to strengthen the ecosystem, improve resource management, and work towards the conservation of marine life. This will draw attention to the urgency of protecting marine life which has been in rapid decline and facing extinction as a result of destructive human acts.

The outcome of the study is targeted towards policymakers, governmental officials, and environmental organizations, as they could use our results to determine the next steps and to guide policies towards improving  the state of our environment. Scientists may also use our models to enhance their research on how climate change affects marine ecosystems. 

We plan on relying on global data that measure geographic distribution of mangroves over time as well as data on oceanic and atmospheric temperature, sea salinity, and sea level (both halosteric- a decrease in sea level caused by the increase in the salinity, and thermosteric- an increase in sea level caused by the melting of ice sheets). Most of the data we will be sourcing is available to the public via the United Nations Environment Programme (UNEP) and The National Centers for Environmental Information (NOAA). The UNEP has compiled a large dataset covering the global geographic distribution of mangroves within the years 1996 - 2016. The UNEP dataset is a \code{.shp} file containing more than 500,000 entries per year, and the NOAA provides us with Oceanic Climate Data Records (CDR) that measure Essential Climate Variables (ECVs) across the globe on a yearly basis.	

Data processing will solve problems involving time gaps in the mangrove data and combine the two datasets (UNEP mangrove dataset and NOAA Oceanic CDR). The temporal resolution of the two datasets is large (only one recording per year), and can not be upsampled reliably. Thus, the approach to studying the datasets will be using model regression as opposed to time-series forecasting.

\newpage

\section{Phase 1: Data Pre-processing} 

\subsection{Dataset 1: Mangrove Distribution}

This part was the most difficult of this phase of the project for us. First, mangrove data was gathered from the \href{https://data.unep-wcmc.org/datasets/45}{UNEP Ocean Viewer website} which contains data on a variety of oceanic metrics. The mangrove distribution dataset was in the form of shapefiles (\code{.shp}), and so to process the data, it was first imported into QGIS, a free and open source geographic information system. The mangrove distribution data spans 7 years (1996, 2007-2010, 2015-2016). To process the data, we used a processing tool provided by QGIS that takes 2 layers (consisting of 2 shape files for different years) and joining the attributes (area from each year) by their location. This took over 15 hours for just two layers and gave rise to 2 problems: first, it was binding in a left join, and thus was missing values unique to the other layer, and second, if a value was not found for the second layer, the value from the first is duplicated. Because of the difficulties faced, we instead generated a grid over the shape file, and found the mangrove distribution within each. This processing comprised the following steps: 
\begin{itemize}        
    \item Using a built in function to fix the geometry data since it contained invalid geometries (such as overlapping shapes) 
    
    \item Generating a grid where each cell is $10000km^2$, bounded by the regions where mangroves are available $(-175.339555556, 179.979555556, -38.856666667, 33.799333333)$
    
    \item Intersecting the grid with each Mangrove shape file such that each shape contains attributes on the grid in which it is contained.
    
    \item Adding geometric data (area and perimeter) for each shape - Exporting the attribute table of each shape file as a \code{.csv} file. The columns would now be: longitude and latitude coordinates, as well as area and perimeter of mangrove population at those coordinates. 
    
    \item The Data was then loaded into a python notebook to be processed further, the columns of the each yearly  dataframe are listed below:
    \begin{table}[h]
    \centering
    \begin{tabular}{ll} \hline
    \textbf{Dataframe Columns}                           & \textbf{Type} \\\hline
    \code{fid}, \code{ogc}, \code{pxlval}, \code{fid\_2} & Leftover from QGIS processing \\\hline
    \code{id}                                            & Grid Cell ID                  \\\hline
    \code{left}                                          & Grid Cell Left Bound          \\\hline
    \code{top}                                           & Grid Cell Top Bound           \\\hline
    \code{bottom}                                        & Grid Cell Bottom Bound        \\\hline
    \code{right}                                         & Grid Cell Right Bound         \\\hline
    \code{area}                                          & Shape Area                    \\\hline
    \code{perimeter}                                     & Perimeter                     \\\hline
    \end{tabular}
    \end{table}

    \item Firstly, columns \code{fid}, \code{ogc\_fid}, \code{pxlval}, and \code{fid\_2} were unneeded since they were leftover from QGIS processing so they were removed. The data was then grouped by the grid cells (defined by the columns: \code{id} (will be renamed to \code{cell\_id}), \code{left}, \code{top}, \code{right}, \code{bottom}), such that each row represents a cell (with its unique id and bounds), instead of a shape, along with the sum of the areas and perimeter of all shapes that fall into this cell. Since the unique grid cells need to be kept track of over the years the data will also need to be processed in such a way that each same grid cell across different years will be given the same identifier, along with adding in grid cells that are present in some years and not in others.
    
    \item The data was separated by year, and so we finally formatted the data frame concatenating the years into one data frame. The columns are now: 
    
    \item Since the data available spans only the following years: 1997, 2007-2010, and 2015-2016, we decided to interpolate the data for the missing years. Note that the method used for interpolation was linear, because of supporting studies that show that mangroves have been declining at approximately a linear rate \cite{Carugati, Spalding, Thomas}.
\end{itemize}
\newpage

\noindent The dataframe at the end of this stage looks like this:
\begin{table}[h]
    \centering
    \begin{tabular}{ll} \hline
    \textbf{Dataframe Columns}                           & \textbf{Type}                 \\\hline
    \code{cell\_id}                                      & Grid Cell ID                  \\\hline
    \code{year}                                          & Year                          \\\hline
    \code{left}                                          & Grid Cell Left Bound          \\\hline
    \code{top}                                           & Grid Cell Top Bound           \\\hline
    \code{bottom}                                        & Grid Cell Bottom Bound        \\\hline
    \code{right}                                         & Grid Cell Right Bound         \\\hline
    \code{area}                                          & Shape Area                    \\\hline
    \code{perimeter}                                     & Perimeter                     \\\hline
    \end{tabular}
\end{table}

\noindent Note: The Coordinate Reference System (CSR) EPSG:4326 - WGS 84 was being worked with during QGIS processing.

\subsection{Dataset 2: NOAA Oceanic Climate Data Records (CDR)}

The National Centers for Environmental Information (NOAA) created the \href{https://www.ncdc.noaa.gov/cdr/oceanic}{Oceanic Climate Data Records} (CDR) to address the need for data on oceanic metrics to better plan against climate change. The data the CDR studies are known as Essential Climate Variables (ECVs) and some of these ECVs are going to be used in this project. The ECV data is in the form of a netCDF format which allows you to query metrics based on time and space. In this case the ECV data is temporally yearly and spatially split on a 1° longitude by 1° latitude grid, worldwide. The ECVs we were concerned with are noted below:

\begin{table}[h]
\centering
\begin{tabular}{ll} \hline
\textbf{Essential Climate Variables (ECV)} & \textbf{Unit/Scale} \\\hline
Heat Content                               & $10^22$ Joules \\\hline
Salinity                                   & Unitless       \\\hline
Temperature                                & ${\circ}$C     \\\hline
Thermosteric Sea Level                     & mm             \\\hline
Halosteric Sea Level                       & mm             \\\hline
Total Steric Sea Level                     & mm             \\\hline
\end{tabular}
\end{table}

The data was loaded in an \code{xarray.DataSet} Object since it is a 4-dimensional dataset (time, depth, lat, lon), while dataframes are only 2-dimensional. The time dimension was encoded to a \code{dateTime} object, and the depth dimension was dropped since it only had a cardinality of 1. Since measurements (from the oceanic dataset) were taken only in water and not on land, and mangroves grow on shores and so on the boundary of water and land, we had lots of missing values for the rows that fall on land. The data was interpolated to allow locations on land to inherit the metrics of nearby oceanic locations. To do this, grid locations with missing CDR data (oceanic metrics) will be interpolated to get nearby essential climate variable (ECV) attributes (inheriting values from their nearest neighbors). The only remaining NaN values are due to a lack of measurements taken from a boundary near Antarctica. We now intersected the mangrove dataset with the CDR (oceanic metrics) dataset.

\subsection{Dataset Combination: }
Dataset 1, Mangrove Geographic Distribution Over Time, and Dataset 2, NOAA CDR, were combined such that the dataset now represents the area of mangroves in each cell along with ECV data. The merging worked by assigning ECV data that are closest to each $10000km^2$ grid square of the mangrove dataset of the same year. The dataframe at the end of this stage has the following columns: \code{cell\_id}, \code{year}, \code{left}, \code{top}, \code{bottom}, \code{right}, \code{area}, \code{perimeter}, \code{heat\_content}, \code{salinity}, \code{temperature}, \code{thermosteric\_sea\_level}, \code{halosteric\_sea\_level}, code{total\_steric}.
\newpage

\section{Phase 2: Data Preparation}

\subsection{Feature Engineering and Feature Crossing}
In order to prepare the data for regression, lag features were added (using the sliding window method). Thus, a new column, 'area-1' was added to the dataset, indicating the area of mangroves at that grid in the following year. 

We also investigated date-time features. With upsampling, we could interpolate to achieve monthly data, and then study the variation of mangrove areas with the seasons or months. However, we do not have thorough knowledge on the rate and trend of area of mangroves as seasons change or throughout months of the year. Because of that, we proceeded with the yearly frequency. 

Finally, since all grids are of the same size, left/right, and top/bottom features are redundant and can be described via a latitude and longitude feature denoting the coordinates at the center of each grid.

\subsection{Statistical EDA}
\begin{figure}[h!]
    \centering
    \includegraphics[width=14cm]{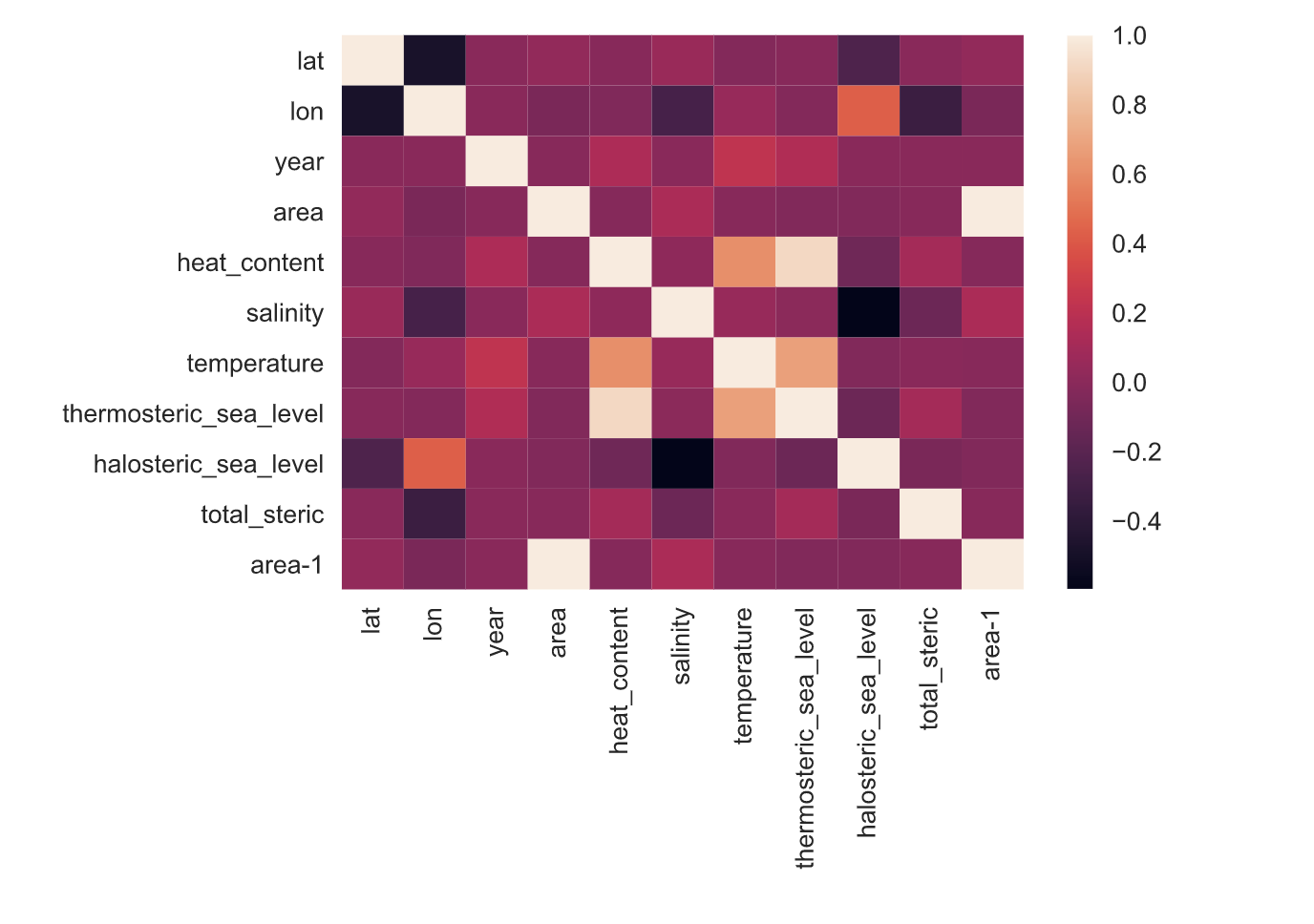}
    \caption{Heat map of the variables in the mangrove dataset}
    \label{fig:p1}
\end{figure}

The correlation matrix (Figure 1) shows that the lagged area value (area-1) is highly correlated with the current area value (area). As there is only slight variation in the area per year, area and area-1 are highly correlated. The second highest correlated feature, according to the figure, is salinity. This confirms the literature review, in which salinity (and temperature) were speculated to be the highest predictors of geographic distribution of mangroves. Also, we can see high correlation between temperature, heat content, and thermosteric sea level, as well as a high correlation between salinity and halosteric sea level. 

\begin{figure}[h!]
    \centering
    \includegraphics[width=14cm]{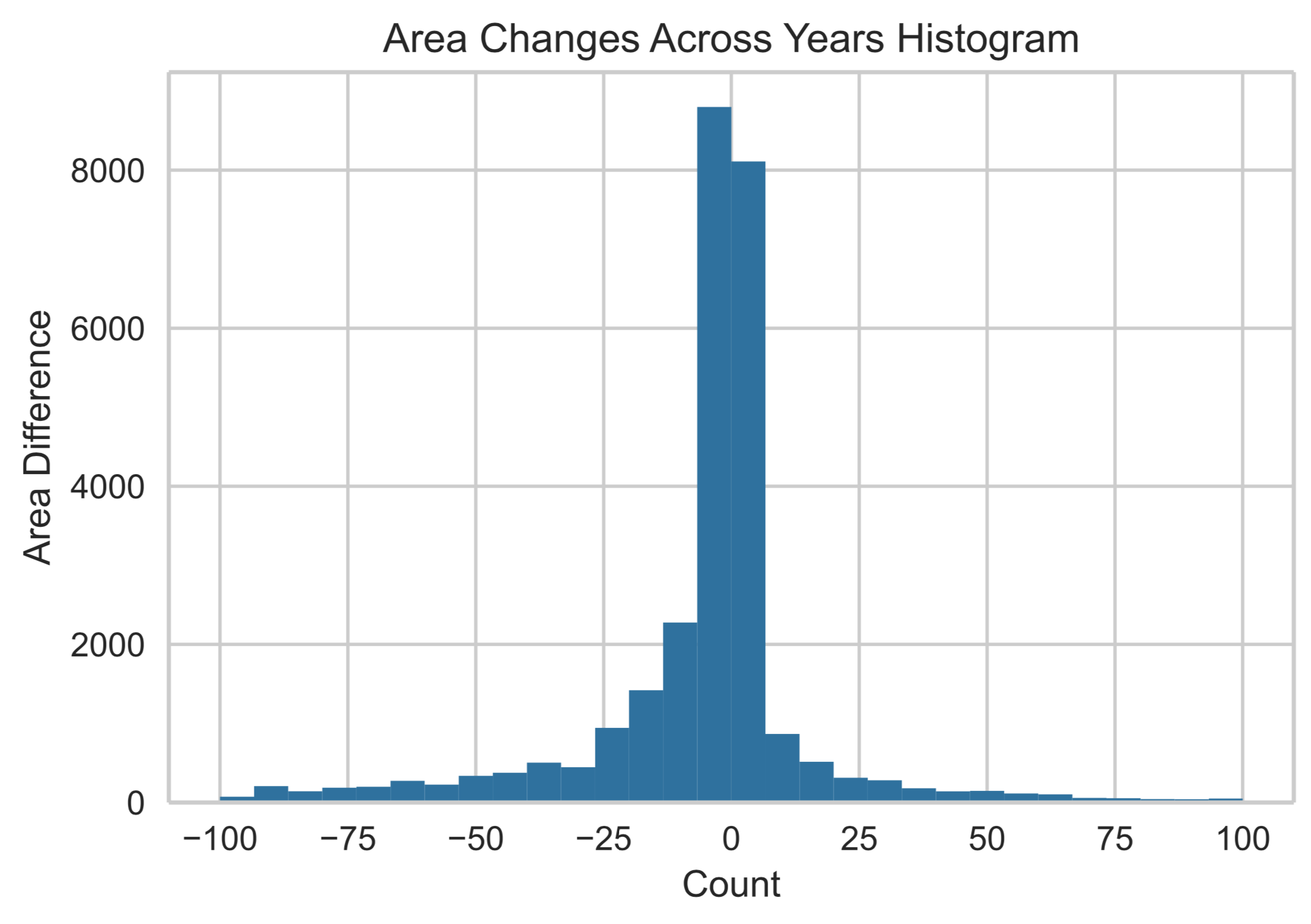}
    \caption{Histogram of the difference between the area at a specific year and the previous year, across all grid locations}
    \label{fig:p3}
\end{figure}

To visualize the change in mangrove area from one year to the next, the histogram (Figure 2) displays the frequencies of the difference in area across all the grids across all consecutive years. The figure shows that most points have a near-zero difference in area from one year to the next. However, the distribution of values on the negative side of the x-axis is higher than that on the positive side. This shows that most values do not change much, however the majority decrease as opposed to increase in their area. Fewer values change significantly, but of those values, the majority decrease while only few show increases in mangrove area. 

Further exploration included plotting a heatmap of heat content on the world map. As suspected, the results show that shore areas on the Atlantic and Pacific Oceans show the highest heat content. 

The exploration also consisted of plotting the distributions of each of the features. The results showed that all the predictors: heat content, salinity, temperature, thermosteric sea level, halosteric sea level, and total steric sea level, appear to be Gaussian. However, the distribution of area shows the highest values near zero, and less and less values at higher ranges. This is expected, since mangrove forests are globally rare, and so mangroves exists only in small quantities in areas besides mangrove forests.

\subsection{Data Cleaning}
To manage outliers, we proceeded to use the interquartile range method, with a factor of 3. This identified around 500 or less outliers for each variable, all less than 0.02\% of the total number of rows. However, 2559 outliers values were identified based on the area of mangroves. [We will plot these outlier values on the map, to study whether they are likely to be erroneous data or true outliers.]

Also, for the ECV variables, we found that none of the variables have zero variance, and so omitted none.

\subsection{Data Transforms}
To magnify the differences in the area, we transformed the scale of the area column. Note that a degree of longitude or latitude is equivalent to approximately 111 Kilometers. The majority of the areas in the dataset are almost zero, and so it was decided to transform these values up $10^5$ orders of magnitude so that they represent values nearer to meters rather than kilometers.

Also in order to prevent the model from considering some columns with a higher weight than others, and to prepare the data for the model, we standardized the data using \code{StandardScaler}. This will also allow us to perform aggregations across columns. The \code{MinMaxScaler} to normalize was not used so as not to squash outlier values.

\subsection{Feature Selection}

\begin{info} 
At this point, data was divided into X (predictor columns), and y (the output, area at the next year). The rows were also divided into training and testing rows (note that this was done randomly, as the problem was framed as a regression, and not a time-series problem). The training data comprised 80\% of the full dataset, and the testing the remaining 20\%.

\end{info}

Recall that unsupervised selection to remove redundant variables has already been done: during data cleaning, we removed redundant variables, such as those with low variance. Additionally, predictor variables were selected specifically from domain research of relevant variables to mangrove distributions. 
Recursive feature elimination (RFE), a wrapper-type feature selection algorithm, was implemented to reveal the features that are most relevant and prune those that are not. The RFE ranked the features in the below order: latitude, longitude, area, and then the remaining 6 at different orders depending on the number of features chosen. Latitude and longitude variables reveal geographic location information, and area reveals the area of the year previous to that being predicted. According to domain research, the remaining variables are correlated (temperature, heat content, and thermosteric sea level are correlated, and salinity and halosteric sea level are correlated). For small numbers of chosen features (n\_features\_to\_select = 5,6,7), the RFE chose one from each of the variables representing salinity, and then one feature representing temperature. For higher levels (n\_features\_to\_select = 8,9), it chose one variable correlated with temperature, followed by salinity.

\subsection{Dimensionality Reduction}

We used principal component analysis (PCA) as a method of dimensionality reduction. There may be high correlation between covariates, and so PCA can produce linear combinations on those that are uncorrelated. A scree plot was developed to reveal the optimal number of principal components (elbow of the plot representing the proportion of variance explained at various principal components). The scree plot reveals multiple elbows, so a different approach was considered. 

The threshold of variance approach revealed that principal components: PC1, 2, 3, and 4 explain approximately 69\% of the variance in the data. Thus, this reduces the dimensionality of the data to 4 components.  

Plotting the results revealed the eigen vectors contributing to PC1 and PC2. It seems the variation in the first principal component was due primarily to the two variables: salinity, and halosteric sea level, and the second was due to temperature, heat content, and thermosteric sea level. This makes sense as halosteric sea level is the increase in the density of the ocean basin, caused by the increase in salinity. On the other hand, thermosteric sea level is that of the decrease in density caused by the melting of ice sheets, which is in turn caused by increasing temperature in the atmosphere and an increase of heat content in the seas.

\subsection{Pattern Analysis}

The K-means clustering algorithm is used to cluster groups into a certain number of clusters, and can reveal patterns in the data. For example, it is possible that some location on the map has a cluster of high areas of mangroves (possibly on the Eastern and Western coasts of Africa or the coast from Florida to Argentina). Using KElbowVisualizer, an elbow was found at 6 clusters. After the data was clustered into 6 clusters, however, no clear ones could be visualized. Thus, the data revealed no clusters on the mangrove data, across all the predictor variables. 

\newpage

\section{Phase 3: Modeling}

\subsection{Linear Regression}
As discussed, the problem was phrased as regression problem due to the low temporal resolution. The initial regression model used was a Linear Regression Model, on three different input data: scaled data, RFE (eliminated) data, PCA (reduced) data with repeated k fold with 10 splits and repeated 10 times.

The results of the linear regression model on scaled data were: 

\begin{itemize}
    \item Linear Regression on Scaled Data Results
    
    \begin{table}[!h]
        \centering
        \begin{tabular}{ll}
        \textbf{Score Metrics}       & \textbf{Scores}          \\ \hline
        Mean Squared Error           & 85496.328980700141983    \\ \hline
        Mean Absolute Error          & 126.050556720954333      \\ \hline
        R$^2$                        & 0.999785335707152        \\ \hline
        \end{tabular}
    \end{table}
    
    Since area does not differ much from one year to the next, and the rate of degradation in the past has been estimated at 1-2\% \cite{Carugati, Thomas} area is highly correlated with area-1 (area at the next year). Thus, the model learns to predict the area at the next year very well.

    \item Linear Regression on RFE Data Results (for number of features selected = 5, which obtained best results)
    
    \begin{table}[!h]
        \centering
        \begin{tabular}{ll}
        \textbf{Score Metrics}       & \textbf{Scores}          \\ \hline
        Mean Squared Error           & 86473.775105464228545    \\ \hline
        Mean Absolute Error          & 114.073932101566740      \\ \hline
        R$^2$                        & 0.999782881534164        \\ \hline
        \end{tabular}
    \end{table}
    
    We ran Linear Regression on RFE Data for (number of features selected = 5, 6, 7, 8, 9), where area was included in all. The best results were obtained with 5 features selected which implies that area is not only the most important predictor, but that the other features may be hurting prediction.

    \item Linear Regression on PCA Data 
    
    \begin{table}[!h]
        \centering
        \begin{tabular}{ll}
        \textbf{Score Metrics}       & \textbf{Scores} \\ \hline
        Mean Squared Error           & 359127293.916927158832550 \\ \hline
        Mean Absolute Error          & 9420.739413871389843         \\ \hline
        R$^2$                        & 0.098302728196023       \\ \hline
        \end{tabular}
    \end{table}
    
    Input of PCA data caused the model to perform worse than the other inputs, likely because area information was sacrificed to reduce dimensionality.
\end{itemize}

\begin{figure}[h!]
    \centering
    \includegraphics[width=8cm]{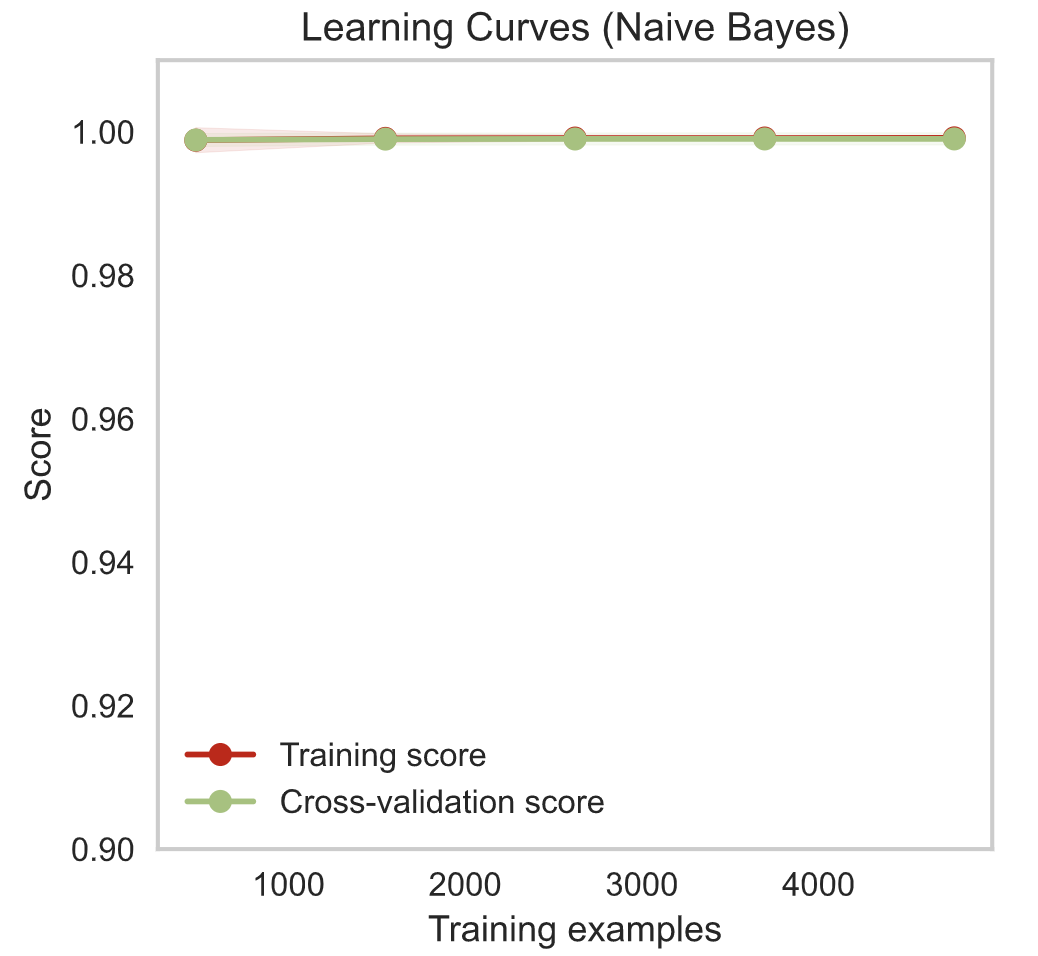}
    \caption{Learning curves of the regression model}
    \label{fig:p4}
\end{figure}
\newpage

It seems since area is highly correlated to the target value area-1, which can be seen in the correlation matrix, and since mangrove area decline over the years is rather predictable (1\% - 2\% changes) the linear regression models learned very accurately. Not only did the model learn accurately but it also learned quickly (refer to the learning curves in Figure 3) which is probably because such a simple near-linear trend was quickly caught by the linear regression model. The curves show high bias and low variance, as the training curve and testing curve are moving together to overlapping. 

\subsection{Support Vector Regression}

The below hyperparameter grid values were chosen, after research on commonly used values. The hyperparameter search was conducted using grid search on commonly used ranges, listed below:
    
    \begin{table}[!h]
        \centering
        \begin{tabular}{ll}
        \textbf{Hyperparameter}       & \textbf{Values}         \\ \hline
        C (regularization parameter)   & 0.1, 10, 100            \\ \hline
        gamma                         & 1, 0.01                 \\ \hline
        epsilon                       & 0.01, 0.0001            \\ \hline
        kernel                        & "linear", "rbt", "poly" \\ \hline
        \end{tabular}
    \end{table}

After performing the regression, the best hyperparameters chosen were:

    \begin{table}[!h]
        \centering
        \begin{tabular}{ll}
        \textbf{Hyperparameter}       & \textbf{Best Values}    \\ \hline
        C (regularization parameter)   & 100                    \\ \hline
        gamma                         & 1                      \\ \hline
        epsilon                       & 0.0001                 \\ \hline
        kernel                        & "linear"                \\ \hline
        \end{tabular}
    \end{table}

\begin{itemize}
    \item SVR on Scaled Data Results
    
    \begin{table}[h!]
        \centering
        \begin{tabular}{ll}
        \textbf{Score Metrics}       & \textbf{Scores}          \\ \hline
        Mean Squared Error           & 74789.932245046657044    \\ \hline
        Mean Absolute Error          & 81.171945695207484       \\ \hline
        R$^2$                        & 0.999812217341856        \\ \hline
        \end{tabular}
    \end{table}

\newpage
    \item SVR on RFE Data Results (for number of features selected = 8, which obtained best results)
    
    \begin{table}[h!]
        \centering
        \begin{tabular}{ll}
        \textbf{Score Metrics}       & \textbf{Scores}         \\ \hline
        Mean Squared Error           & 75084.828552276841947   \\ \hline
        Mean Absolute Error          & 81.451736176114878      \\ \hline
        R$^2$                        & 0.999811476915828       \\ \hline
        \end{tabular}
    \end{table}
    
    \item SVR on PCA Data Approximate Results
    
    \begin{table}[h!]
        \centering
        \begin{tabular}{ll}
        \textbf{Score Metrics}       & \textbf{Scores}              \\ \hline
        Mean Squared Error           &  430489789.088259518146515   \\ \hline
        Mean Absolute Error          &  7418.253368059360582        \\ \hline
        R$^2$                        &  -0.080874316531743          \\ \hline
        \end{tabular}
    \end{table}
    
\end{itemize}

\begin{figure}[h!]
    \centering
    \includegraphics[width=8cm]{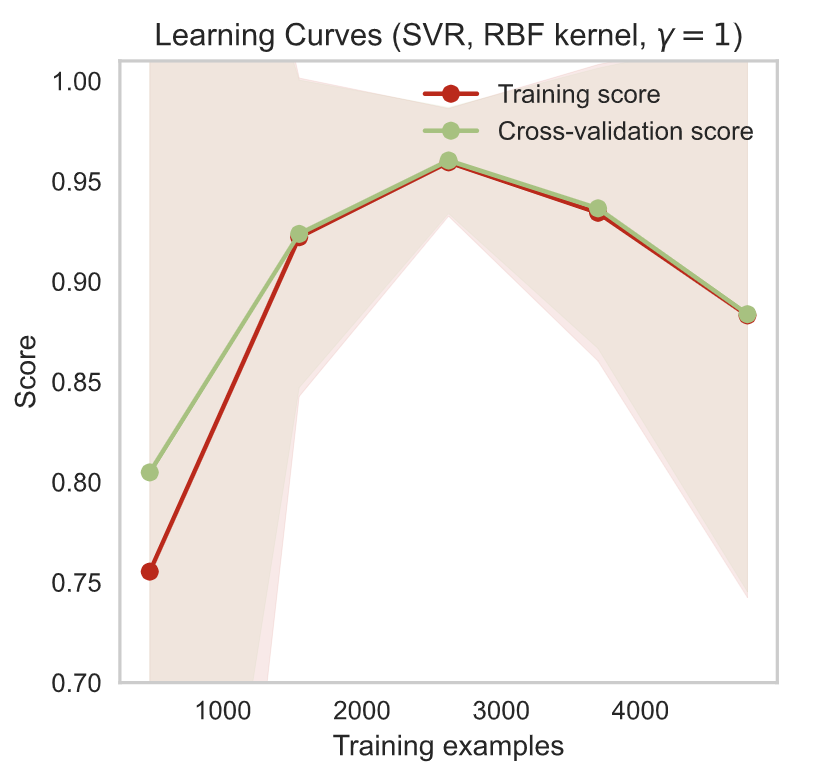}
    \caption{Learning curves of the SVR model}
    \label{fig:p5}
\end{figure}

As the model is fed more and more training examples, the training score improves, as the model is learning and improving its predictions. Variance decreases as the two curves approach each other as they improve their scores. At between 2000 and 3000 examples, the scores begin to decrease. This indicates that the model may be overfitting to the training examples, and is no longer able to generalize. 

\subsection{XGBoost Regressor}

The next model used was an XGBoost model, which led to the following scores:

\begin{itemize}
    \item XGBoost on Scaled Data Results
    
    \begin{table}[!h]
        \centering
        \begin{tabular}{ll}
        \textbf{Score Metrics}       & \textbf{Scores}          \\ \hline
        Mean Squared Error           & 407131.900876009953208    \\ \hline
        Mean Absolute Error          & 135.668502304634131      \\ \hline
        R$^2$                        & 0.998977772699256        \\ \hline
        \end{tabular}
    \end{table}
    
    The model performed much worse than the linear regression and the SVR models, with an r$^2$ value of around -0.08. Thus, the model does not learn to predict the output (area at the following year) given the predictors.
    
\newpage
    
    \item XGBoost on RFE Data Results (for number of features selected = 9, which obtained best results)
    
    \begin{table}[!h]
        \centering
        \begin{tabular}{ll}
        \textbf{Score Metrics}       & \textbf{Scores}          \\ \hline
        Mean Squared Error           & 375950.399429603363387   \\ \hline
        Mean Absolute Error          & 134.546146491572330     \\ \hline
        R$^2$                        & 0.999056063253222        \\ \hline
        \end{tabular}
    \end{table}
    
    Since the total number of features is 11, and the RFE selected only 9 features, it is likely that some features were distorting the model training.
    
    \item XGBoost on PCA Data Results
    
    \begin{table}[!h]
        \centering
        \begin{tabular}{ll}
        \textbf{Score Metrics}       & \textbf{Scores}          \\ \hline
        Mean Squared Error           & 165254454.406062394380569 \\ \hline
        Mean Absolute Error          & 6912.241500903080123     \\ \hline
        R$^2$                        & 0.585078903176126     \\ \hline
        \end{tabular}
    \end{table}
    Here, the performance diminished, likely because area information was reduced by PCA to reduce dimensionality. 
    
\end{itemize}

\begin{figure}[h!]
    \centering
    \includegraphics[width=8cm]{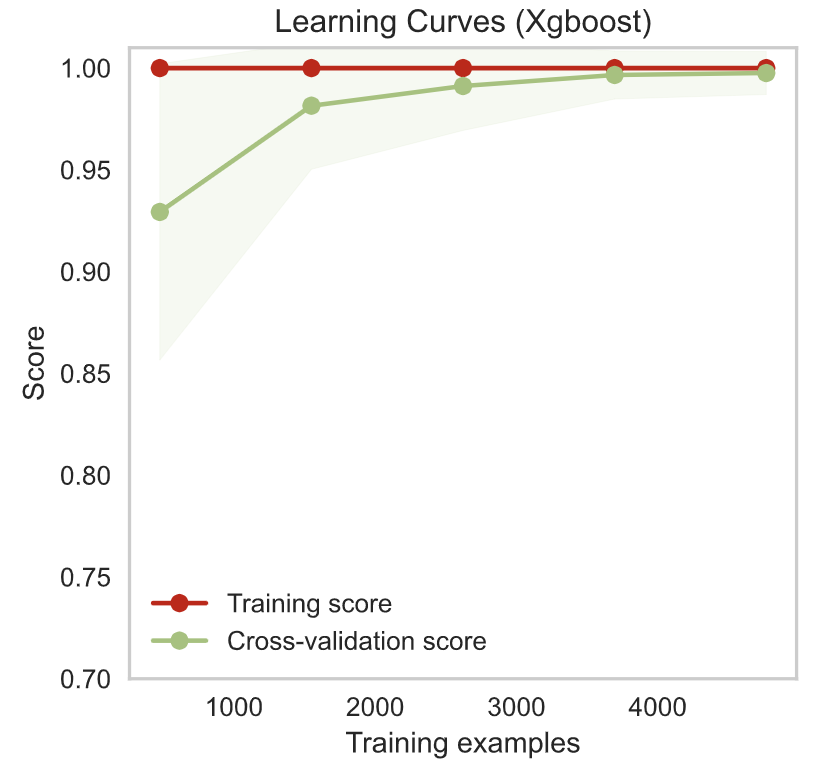}
    \caption{Learning curves of the XGBoost model}
    \label{fig:p6}
\end{figure}

The learning curves for the XGBoost model (Figure 5) show that the training score begins at an almost perfect level from even the lowest number of training examples. This confirms that mangrove distributions are simple to model. With more training, the cross-validation score improves and approaches the training score. This means that the model is learning to generalize. Both scores are high, so the model does not have high bias, and the models are near each other, so the model does not suffer from high variance. 

\subsection{Random Forest Regressor}

\begin{itemize}
    \item Random Forest Regressor on Scaled Data
    \begin{table}[!h]
        \centering
        \begin{tabular}{ll}
        \textbf{Score Metrics}       & \textbf{Scores}          \\ \hline
        Mean Squared Error           & 3983085.323880965355784\\ \hline
        Mean Absolute Error          & 210.522852037009699   \\ \hline
        R$^2$                        & 0.989999264242119    \\ \hline
        \end{tabular}
    \end{table}
    
    \item Random Forest Regressor on RFE Data (number of features selected = 9)
    \begin{table}[!h]
        \centering
        \begin{tabular}{ll}
        \textbf{Score Metrics}       &       \textbf{Scores}          \\ \hline
        Mean Squared Error           & 2106631.284070075955242\\ \hline
        Mean Absolute Error          & 189.624750017495586   \\ \hline
        R$^2$                        & 0.994710667460485     \\ \hline
        \end{tabular}
    \end{table}
    
    \item Random Forest Regressor on PCA Data 
    \begin{table}[!h]
        \centering
        \begin{tabular}{ll}
        \textbf{Score Metrics}       &
        
        \textbf{Scores}          \\ \hline
        Mean Squared Error           & 165254454.406062394380569\\ \hline
        Mean Absolute Error          & 6912.241500903080123   \\ \hline
        R$^2$                        & 0.585078903176126     \\ \hline
        \end{tabular}

    \end{table}
    
\end{itemize}

\begin{figure}[h!]
    \centering
    \includegraphics[width=8cm]{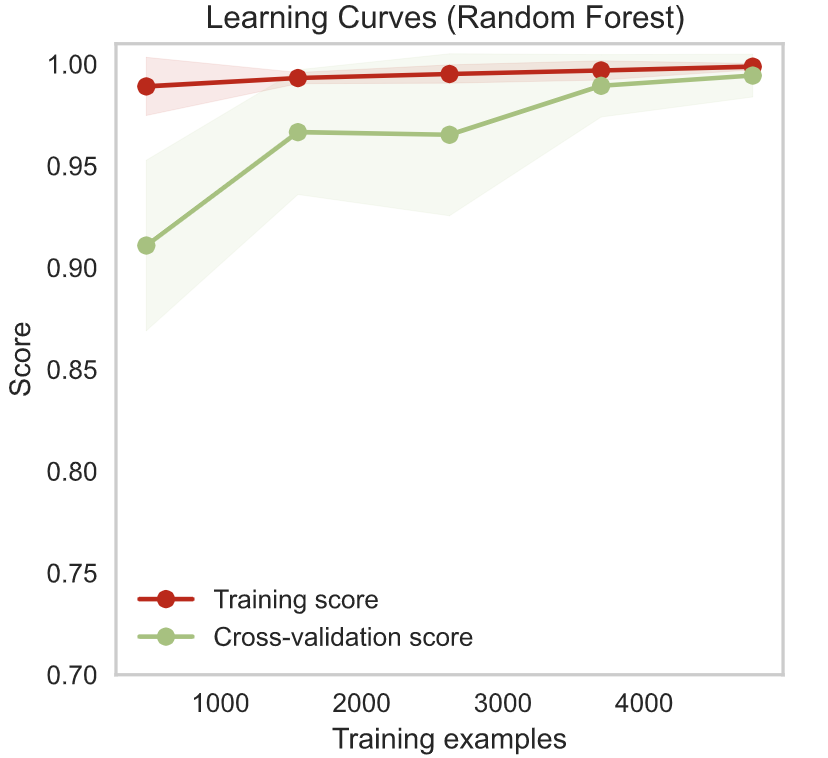}
    \caption{Learning curves of the Random Forest Regressor model}
    \label{fig:p7}
\end{figure}

Similar to the learning curves for the XGBoost model the learning curves for the Random Forest Regressor (Figure 6) show an already high training score that improves slightly with more training. The cross-validation score also improves and approaches the training score. The model is thus learning to generalize. The model is neither suffering from high bias or high variance.  

\newpage

\section{Phase 4: Analyses}

\subsection{Statistical Hypothesis Testing}

We conducted a hypothesis test to investigate whether there exists a significant difference between the scores across different models. The parameter alpha or the significance level, was set to 0.05, meaning that there is less than a 5\% possibility that the results are due to random chance. The hypotheses are listed below:

\begin{itemize}        
    \item Null Hypothesis (H0): There is no significant difference between the scores across models (p $> 0.05$). 
    \item Alternative Hypothesis (H1): There is a significant difference between the scores across models (p $\leq 0.05$).
\end{itemize}

To choose a hypothesis test that is best suited to our experiment, we considered the t-test, pairwise t-test, analysis of variance (ANOVA) test, and the repeated measures ANOVA test. All tests require that the data samples are of a Gaussian distribution, a feature confirmed in the above visualizations since areas are near-Gaussian distributed. Since we wanted to compare the scores across more than two models, we opted for ANOVA instead of t-tests, and since the distributions are independent of one another, we used the ANOVA test and not the repeated measures ANOVA test.

The results of the hypothesis test indicated a significant difference between the scores. Since this significant difference exists, and the SVR model produced the highest score values (significantly higher, based on our hypothesis tests), it is considered the best performing model overall.

\subsection{Statistical Hypothesis Testing Results}

The support vector regressor (SVR) model was the best performing model from those investigated. We conducted further hypothesis tests to compare the model’s performance on scaled data versus data after feature selection. The hypotheses are:

\begin{itemize}        
    \item Null Hypothesis (H0): There is no significant difference between the number of features selected in the scores of the SVR model. (p $> 0.05$).
    \item Alternative Hypothese (H1): There  is a significant difference between the number of features selected in the scores of the SVR model (p $\leq 0.05$).
\end{itemize}

The lists of scores achieved through cross validation (CV) was compared using the below steps:		
\begin{enumerate}
    \item We conducted ANOVA tests to investigate whether a significant difference exists between the feature selection data, i.e. Is there a difference between model results using data where number of features selected = 5, 6, 7, 8, or 9? The results yielded no significant difference.
    \item Thus, we considered the model with the highest r2 score (model with number of features selected = 8). Note that there was no significant difference between the results, so any could be used.
    \item Finally, we conducted a t-test to assess the difference between the model on scaled data versus data after feature selection (where number of features selected = 8). The reason we opted for a t-test instead of the ANOVA test here is because only two models were being compared.
\end{enumerate}

The result showed a failure to reject the null hypothesis. This means that there is no significant difference between the scores within feature selection (for number of features selected = 5, 6, 7, 8, or 9). So, we considered the scores yielded when the number of features selected was 7. We then conducted a t-test since we now had only two distributions to investigate: the scores of the model before feature selection and after feature selection (with number of features = 7). The hypotheses of the t-test are:

\begin{itemize}        
    \item Null Hypothesis (H0): There is no significant difference between scores of the SVR model before and after feature selection. (p $> 0.05$).
    \item Alternative Hypothesis (H1): There  is a significant difference between scores of the SVR model before and after feature selection (p $\leq 0.05$).
\end{itemize}

The results of this t-test indicated a significant difference between the scores of the model before and after feature selection. In effect, feature selection significantly benefited the model’s performance. This shows that using the full number of predictors decreased the model’s performance and eliminating certain features benefited its performance. As previously noted, there exists a correlation between temperature, heat content, and thermosteric sea level, as well as a correlation between salinity and halosteric sea level.Thus, eliminating one of two correlated variables eliminates redundancy. The random feature elimination (RFE- a feature selection method) then provides the benefit of eliminating redundant data, consequently improving model performance.

\subsection{SHAP Interpretability}

SHAP results allow for interpretability, i.e. ’decoding’ the black box of the results. For global interpretabil- ity, we printed two plots: the SHAP bar plot (Figure 7) shows the results of feature importance. As pre- dicted, the area at a specific location is the highest predictor of the area in the next year. Not only is it highly important, but its importance trumps that of the other predictors. The variable with the highest impact on model output magnitude was that of area. The second plot (Figure 8) is a SHAP summary plot, which depicts the impact of each point on the model output. Again, it reveals that area had the most impact. The horizontal location shows whether the effect of each variable is associated with a higher or lower prediction. For area, as predicted, a high area is associated with a high and positive impact on the area at that point in the following year. This would definitely look reasonable to a domain expert being served the model’s predictions. Figure 7 and 8 both indicate that the area is the highest predictor of the area at the next year, which is reasonable because we know, based on the literature review, that mangrove distributions only fluctuate very minimally, and so the salinity, heat content, and other oceanic predictors would not have as big of an impact as the area in predicting the area at the next year. Also, the decision plot for salinity (Figure 9) demonstrates that given the available predictors, salinity interacts most with latitude, and that there is a linear and positive trend between salinity and latitude values. Finally, the decision plot for area (Figure 10) indicates that area interacts most with the longitude coordinate, and that there is a positive trend between area and longitude: the area of mangroves increases as the longitude coordinate increases.	

\begin{figure}[h!]
    \centering
    \includegraphics[width=12cm]{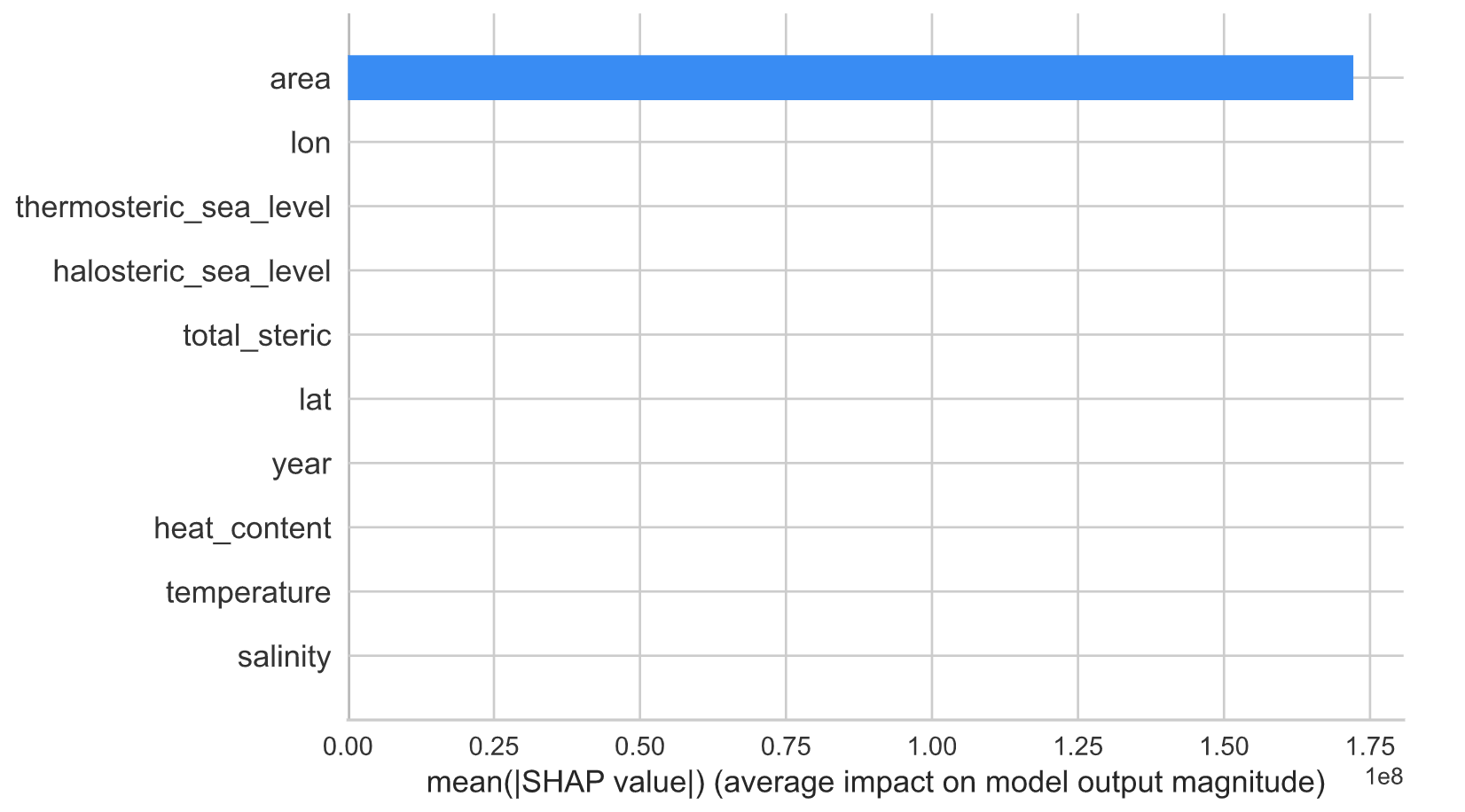}
    \caption{SHAP Bar Plot}
    \label{fig:s1}
\end{figure}

\begin{figure}[h!]
    \centering
    \includegraphics[width=10cm]{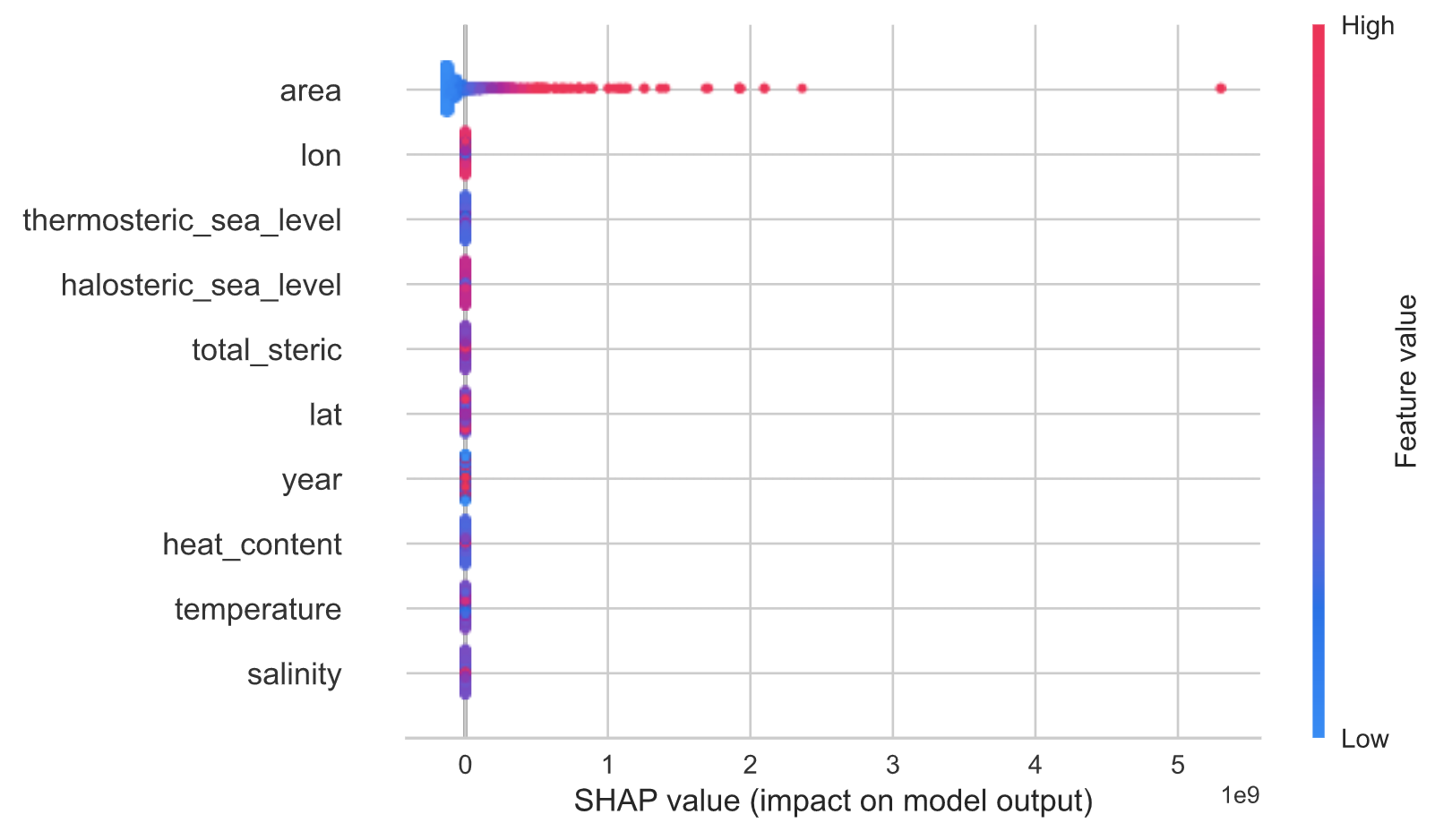}
    \caption{SHAP Summary Plot}
    \label{fig:s2}
\end{figure}
\newpage
\begin{figure}[h!]
    \centering
    \includegraphics[width=10cm]{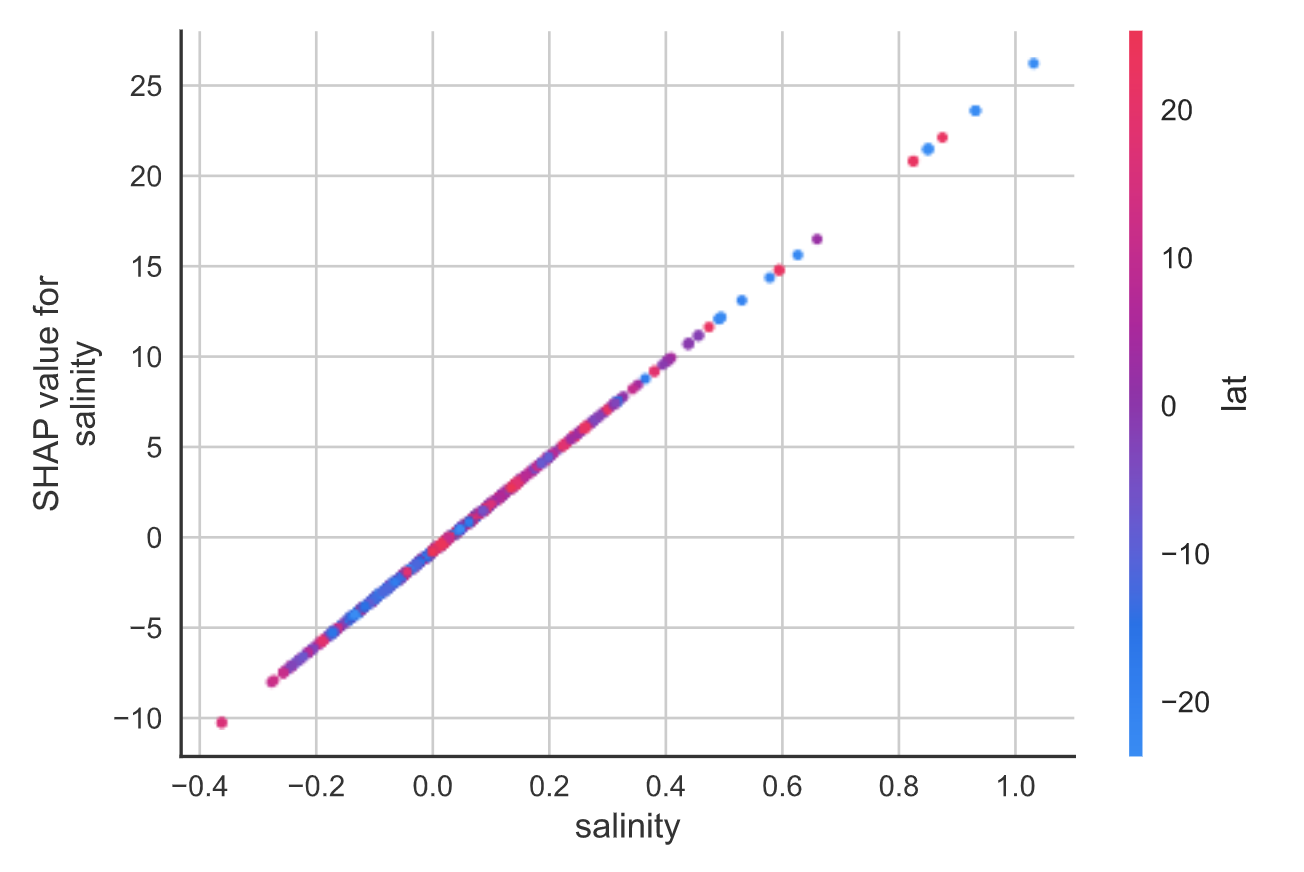}
    \caption{SHAP Dependence Plot for Salinity}
    \label{fig:s3}
\end{figure}

\begin{figure}[h!]
    \centering
    \includegraphics[width=10cm]{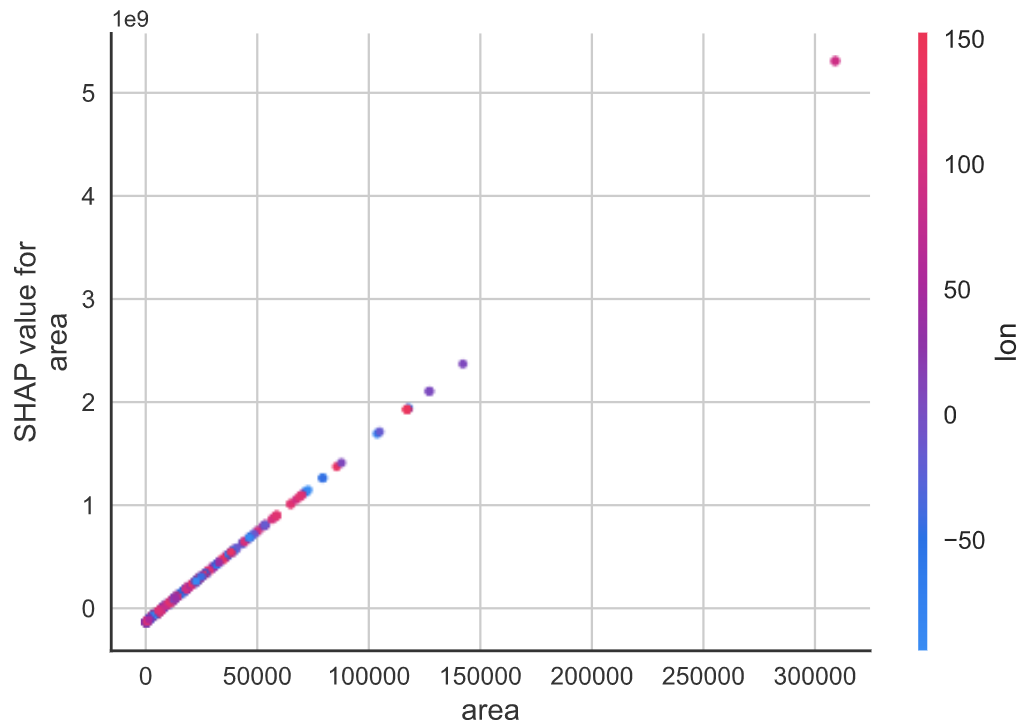}
    \caption{SHAP Dependence Plot for Area}
    \label{fig:s3}
\end{figure}

\subsection{Graduate Work}

\begin{itemize}

\item \textbf{Model Agnostic Meta Learning}

In meta-learning, there is a meta-learner and a learner. The meta-learner trains the learner on a training set that contains a large number of different tasks. In this stage of meta-learning, the model will acquire a prior experience from training and will learn the common features representations of all the tasks.
We tried to apply Model Agnostic Meta Learning (MAML) to our dataset which provides a good initialization of a model's parameters to achieve an optimal fast learning, because MAML promises to overcome some overfitting problems, that we may have been encountering.
On our dataset, MAML proved to be ineffective, considering it resulted in a negative adaption rate. This means that the meta-learner was not able to adapt the learner to do predictions on this dataset.

\item \textbf{Conformalized Quantile Regression}

CQR is a methodology that aims to create a short prediction interval that is locally variable or adaptive on heteroscedastic data without making any distributional assumptions. This provides a probabilistic perspective that tries to improve on accuracy by limiting the amount of miscoverage. This miscoverage rate usually set to 0.1 accounts to allowing the error rate to be no longer than 10\% and with that a full view of how the predict value would vary with a small amount of error. The smaller this variety (length of the predicted interval), the more information we can get about predictions. For this purpose we did CQR on our data using neural network and random forest. The coverage for both models were not satisfied, as both had less 90\% coverage.

This is works same as the non-formalized Quantile regression as it does not give guaranteed coverage but a near coverage and is adaptive to heteroscedastic data. Although this is the case, we still did this on our data to predict an interval for the area of mangroves on the year 2017 given that we already have the data for the year 2016.
A note to be mentioned that while CQR tries to control the amount of miscoverage, meta learning allows for the miscoverage to increase and be more variable. Hence both contradict each other and it is better for them not to be applied together.

\item \textbf{Time Series Forecasting}

The data that was being worked with had elements of time series data but varied from traditional time series data because it consisted of multiple series, one for each grid cell - denoted by the spatial longitude, latitude components. This is why traditional time series forecasting techniques like ARIMA, RNN, and LSTMs was not applicable to the data and required regression models, but certain time series data techniques did come in very useful. Data interpolation was used to fill in the data for the missing years and lagged area values to feature engineer the target variable.

\end{itemize}

\subsection{Achievements and Limitations}

The results of the analyses confirm that the general trend of mangrove distribution is decreasing at a near linear rate with time. The results also allow for the prediction of the mangrove distribution at a specific grid location on the globe using the ‘previous area’ and the ‘climate’ variables: heat content, temperature, sea salinity, and the various sea levels. However, this study is not without its limitations; to be sure, we encountered some difficulties that could potentially be addressed with better data. The work has also revealed new avenues for future research. 

\subsubsection{Limitations}
One of the limitations we encountered was due to low temporal resolution (recall that low temporal resolution here means having only one recording for mangrove distribution each year). With low resolution, the model can not account for the changes that occur from one year to the next, reducing the accuracy of the prediction. To improve results and achieve more accurate predictions, it is ideal to get access to data with higher temporal resolution, for example recordings each month as opposed to each year. This would allow for more training data as well as more continuous data. 

The model would also benefit from having more data in general. The current model only has data for the years 2007-2010 and 2015-2016, and thus a big chunk of the data (4 years of mangrove distributions) had to be interpolated. Although the data that was interpolated only covered four years, this may have adversely affected the results. Interpolating data in such a way may have caused the trend to appear linear not because the trend itself is truly linear, but rather because the method of the interpolation was linear. One further limitation comes from the fact that we trained on a variety of models, including complex models.The complex models easily overfit to the fairly simple trend of mangrove decline (an almost linear decline).

\subsubsection{Next Steps}
Moving forward, the model can be applied to similar data on other plant species (including coral reef populations, kelp, etc). Not only could the distributions of these populations be predicted, but using similar analyses, the extent of symbiotic and asymbiotic relationships between species could also be revealed. 		

\subsubsection{SHAP Interpretability Results}
The SHAP interpretability results do promote trust from the part of the end user in the sense that the decline of mangrove populations in different areas is minimal, and so the biggest predictor would be the area at the previous year. However, the figures show almost no results for the remaining predictors. This may not promote trust from the users, who may wonder if any effect of these predictors exists at all and if so, what they are. Further research could reformulate the question and attempt to use solely oceanic statistics in predicting the effect of these oceanic statistics on the distribution of mangroves (and other organisms). 

\subsubsection{Using 2016 Data to Predict Mangrove Areas for 2017}
Using the best performing model (SVR) and data from the year 2016 (global mangrove distribution in 2016 and ocean statistics), the mangrove distribution for the year 2017 was predicted. The results showed an average decrease in mangrove area of $11.7x10^-5$ degrees squared. This demonstrates that on average, the distribution in all grid locations is decreasing. Negative areas in grid locations indicate that some mangroves will cease to exist entirely. What is alarming is that the rate of decline is overwhelming and consistent, indicating that at this rate, mangroves are highly at risk of decline and eventual extinction. This justifies further research in the area and further conservation efforts.

\newpage
\bibliographystyle{unsrt}

\begin{thebibliography}{9}
\bibitem{Carugati} Carugati, L., Gatto, B., Rastelli, E. et al. Impact of mangrove forests degradation on biodiversity and ecosystem functioning. Sci Rep 8, 13298 (2018). https://doi.org/10.1038/s41598-018-31683-0

\bibitem{FAO} FAO. 2020. The State of World Fisheries and Aquaculture 2020. Sustainability in action. Rome. https://doi.org/10.4060/ca9229en

\bibitem{Mendelsohn} Mendelsohn, Robert. 2011. “The Impact of Climate Change on Land”. Climate Change and Land Policies, eds. Ingram, Gregory K. and Yu-Hung Hong. Cambridge, MA: Lincoln Institute of Land Policy.

\bibitem{Reed} Reed D.H. (2012) Impact of Climate Change on Biodiversity. In: Chen WY., Seiner J., Suzuki T., Lackner M. (eds) Handbook of Climate Change Mitigation. Springer, New York, NY. https://doi.org/10.1007/978-1-4419-7991-9\_15

\bibitem{Spalding} Spalding, M.D. et al. 2001b. The Global Distribution and Status of Seagrass Ecosystems. Cambridge, UK: World Conservation Monitoring Centre

\bibitem{Sutariya} Sutariya, Sachin \& Ankur, Hirapara \& Meherbanali, Momin. (2020). Impact of Climate Change on Agriculture. 

\bibitem{Thomas} Thomas N, Lucas R, Bunting P, Hardy A, Rosenqvist A, Simard M (2017) Distribution and drivers of global mangrove forest change, 1996–2010. PLoS ONE 12(6): e0179302. https://doi.org/10.1371/journal.pone.0179302
\end{thebibliography}


\end{document}